\documentclass[lettersize,journal]{IEEEtran}
\usepackage{amsmath,amsfonts}
\usepackage{booktabs}
\usepackage{algorithmic}
\usepackage{algorithm}
\usepackage{array}
\usepackage[caption=false,font=normalsize,labelfont=sf,textfont=sf]{subfig}
\usepackage{textcomp}
\usepackage{stfloats}
\usepackage{url}
\usepackage{verbatim}
\usepackage{graphicx}
\usepackage{multirow}
\usepackage{makecell}
\usepackage{cite}
\begin{document}

\title{PRIMED: Adaptive Modality Suppression for Referring \\ Audio-Visual Segmentation via Biased Competition}

\author{
Yuchen He,
Jing Zhang$^{\ast}$%
\thanks{Yuchen He and Jing Zhang are with the School of Information Science and Engineering, East China University of Science and Technology, Shanghai, China. (email: heyuchen@mail.ecust.edu.cn; jingzhang@ecust.edu.cn)}
\thanks{*Corresponding author: Jing Zhang.}
}
        % <-this % stops a space



\maketitle

\begin{abstract}
Referring Audio-Visual Segmentation (Ref-AVS) seeks to localize and segment target objects in video frames based on visual, auditory, and textual referring cues. The task is challenging because the relevance of different modalities varies across referring expressions and scenes, while existing methods typically treat multimodal cues as homogeneous inputs for fusion, prompting, or reasoning, making them vulnerable to irrelevant or misleading modalities. To address this problem, we propose PRIMED, inspired by the biased competition theory in cognitive neuroscience, which explicitly models both visual perception and language-driven prior modulation, and enables more accurate Ref-AVS by adaptive modality suppression. Specifically, a Modality Prior Decoder first estimates whether the referring expression relies primarily on audio, vision, or their joint interaction, generating a modality prior to adaptively guide high-level attention. A Token Distiller further extracts compact global visual tokens from high-level features and shares them across Competition-aware Cross-modal Fusion modules to provide hierarchical global context. Additionally, we introduce a Spatial-Aware Semantic Alignment loss to further enhance foreground-background discrimination through contrastive learning. Extensive experiments on the Ref-AVS benchmark demonstrate that PRIMED achieves state-of-the-art overall performance.
\end{abstract}


\section{Introduction}
\IEEEPARstart{R}{eferring} Audio-Visual Segmentation (Ref-AVS) aims to segment target objects in video frames based on visual, auditory, and textual cues. Recent advances in multimodal learning have achieved notable success in bimodal tasks such as text-guided visual grounding and audio-visual understanding~\cite{gavrilyuk2018actor, tian2018audio}, but extending these approaches to Ref-AVS remains highly challenging due to the need for fine-grained cross-modal alignment across three heterogeneous modalities. Compared with traditional Referring Video Object Segmentation (R-VOS)~\cite{khoreva2018video, seo2020urvos, li2023robust}, which relies solely on visual and linguistic inputs, Ref-AVS introduces an additional auditory modality that provides complementary cues while simultaneously escalating the complexity of cross-modal reasoning. Driven by its capacity to handle such complex real-world signals, Ref-AVS has attracted growing attention due to its potential in applications such as intelligent video editing~\cite{zhang2024effived, yoon2025raccoon} and embodied robotics~\cite{fang2024embodied, huang2025roboground}.

Existing Ref-AVS approaches can be broadly categorized into three paradigms based on how multimodal information is utilized: fusion-based, prompt-based, and reasoning-based methods. Fusion-based methods, such as EEMC~\cite{wang2024ref} and VoCa~\cite{guo2025leveraging}, mainly focus on unified multimodal representations and strengthen spatio-temporal modeling for segmentation. Prompt-based methods~\cite{wang2025sam2, radman2025tsam} reformulate multimodal inputs as prompts to guide SAM/SAM2 mask decoders, significantly improving generalization and mask quality. In contrast, unified audio-visual scene understanding models such as Crab~\cite{du2025crab} introduce explicit cooperation and reasoning-enhanced instruction tuning, while Ref-AVS-oriented methods such as AURORA~\cite{luo2025aurora} and TGS-Agent~\cite{zhou2025think} further incorporate structured reasoning or task decomposition to better handle complex audio-visual scenes. Despite their differences, these approaches share a common limitation: they primarily treat multimodal inputs as information to be fused, prompted, or reasoned over, without explicitly modeling the distinct roles and varying relevance of each modality in guiding target selection. As a result, they may struggle when multiple candidate regions partially satisfy the referring expression, since irrelevant or misleading modalities may interfere with robust target segmentation.

\begin{figure}[t]
  \centering
  \includegraphics[width=\linewidth]{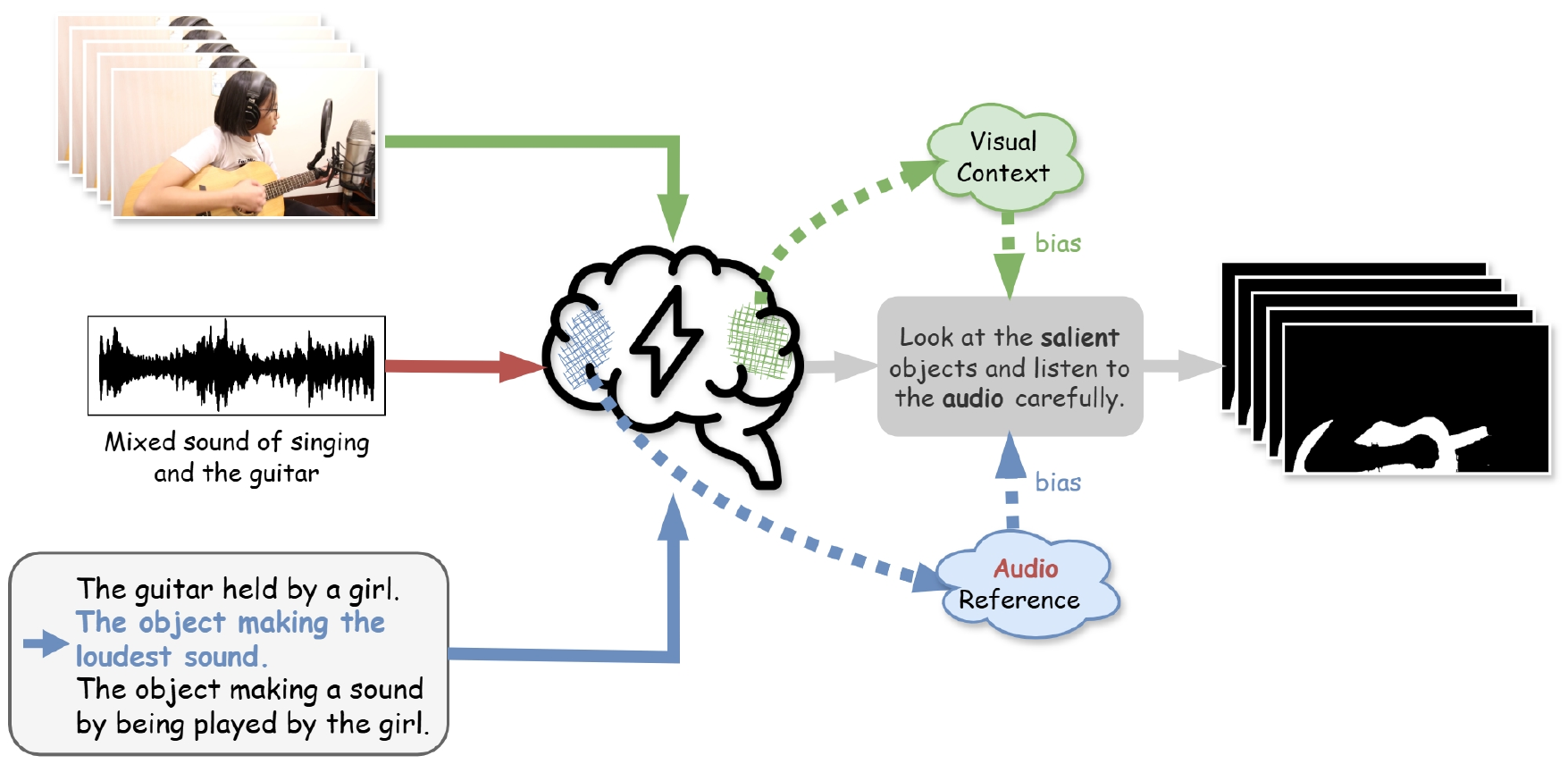}
  \caption{In cognitive neuroscience, multiple candidate regions compete for limited representation, namely biased competition. In Ref-AVS, the competition is under the joint influence of visual context and the task-driven goal from the referring expression.}
  \label{fig:motivation}
\end{figure}

Biased competition theory in cognitive neuroscience~\cite{desimone1995neural, beck2009top} suggests that multiple stimuli compete for limited representational capacity, while bottom-up saliency and top-down task signals bias this competition toward task-relevant targets. As a result, relevant regions are enhanced and distractors are suppressed. This perspective naturally aligns with Ref-AVS, where multiple candidate regions may partially match the referring expression across different modalities, requiring the model to resolve their competition under visual, auditory, and linguistic constraints. In this context, bottom-up evidence arises from hierarchical visual features, whereas the referring expression provides top-down guidance by specifying the target conditions and providing an initial cue about modality relevance. The whole procedure is illustrated in Fig.~\ref{fig:motivation}.

Motivated by this observation, we explicitly formulate Ref-AVS as a competition-aware segmentation process, where bottom-up multi-scale visual evidence is preserved and complemented by global visual context, while the top-down modality prior biases the selection of target regions at the highest visual stage.

Building upon the above insight of competition-aware segmentation with the top-down modality prior, we propose PRIMED (\textbf{PR}ior-\textbf{I}nformed \textbf{M}odulation and \textbf{E}mbedding \textbf{D}istillation), a novel SAM2-based framework that explicitly models biased competition in Ref-AVS. Specifically, we first construct a unified multimodal semantic representation by aligning textual and auditory cues, and introduce a lightweight Modality Prior Decoder (MPD) to generate the language-guided modality prior, which provides a top-down signal for target-aware modulation. To model competition across the visual hierarchy, we design a cascade of Cross-modal Biased Competition Fusion (CBCF) modules that integrate multimodal semantics into multi-scale visual features. In particular, the modality prior is applied at the highest-level features to modulate global visual perception, while lower-level features are guided by semantic interactions for fine-grained alignment. Meanwhile, we distill compact global visual tokens from high-level visual embeddings and propagate them to all CBCF modules, enabling consistent global context across different stages. Finally, we introduce a Spatial-Aware Semantic Alignment (SASA) loss to enhance foreground-background discrimination by aligning multimodal semantics with target regions while separating them from the background. In summary, our contributions are threefold:

\begin{enumerate}

\item We revisit Ref-AVS from the perspective of \textbf{modality-aware biased competition}, highlighting the importance of modeling the distinct roles and varying relevance of multimodal cues in guiding target selection.

\item We propose \textbf{PRIMED}, a novel SAM2-based framework that explicitly models \textbf{competition-aware multimodal modulation}. Specifically, the language-guided \textbf{Modality Prior Decoder} provides a top-down modality prior, while Cross-modal Biased Competition Fusion modules progressively integrate multimodal semantics into hierarchical visual features to bias candidate competition.

\item We further introduce a \textbf{Token Distiller} to propagate compact global visual context across the hierarchy, and a \textbf{Spatial-Aware Semantic Alignment} loss to enhance foreground-background discrimination, leading to improved segmentation accuracy and robustness under modality ambiguity.

\end{enumerate}

\section{Related Work}
\subsection{Audio-Visual Segmentation}
Audio-Visual Segmentation (AVS)~\cite{zhou2022audio, zhou2025audio} aims to generate pixel-level masks for sounding objects in unconstrained videos. Existing approaches predominantly adopt a query-based paradigm, where audio signals are treated as query tokens to guide the localization of sounding regions. For instance, methods such as CATR~\cite{li2023catr}, AVSegFormer~\cite{gao2024avsegformer}, and TransAVS~\cite{ling2024transavs} integrate audio queries into transformer-based decoders to model cross-modal interactions. Subsequent works, including AQFormer~\cite{huang2023discovering} and CQFormer~\cite{lv2025consistency}, further enhance this paradigm by designing audio-conditioned or consistency-aware queries to improve cross-modal alignment and robustness in dynamic scenes.

To address noisy or unstable audio signals, QDFormer~\cite{li2024qdformer} introduces quantization-based semantic decomposition for more robust audio-visual representations. In parallel, recent efforts explore adapting visual foundation models to AVS. AV-SAM~\cite{mo2023av} and SAMA-AVS~\cite{liu2024annotation} adapt SAM to AVS through audio-visual fusion or adapter tuning, while GAVS~\cite{wang2024prompting} adopts an encoder-prompt-decoder paradigm that reformulates audio as prompts for SAM-based audio-visual segmentation.

Despite these advances, AVS primarily focuses on sounding-object segmentation under dual-modality supervision and lacks mechanisms for language-guided target disambiguation. As a result, it cannot effectively handle scenarios where multiple candidates exist and modality relevance varies, limiting its applicability to more complex multimodal grounding settings such as Ref-AVS.

\begin{figure*}[t]
  \centering
  \includegraphics[width=\linewidth]{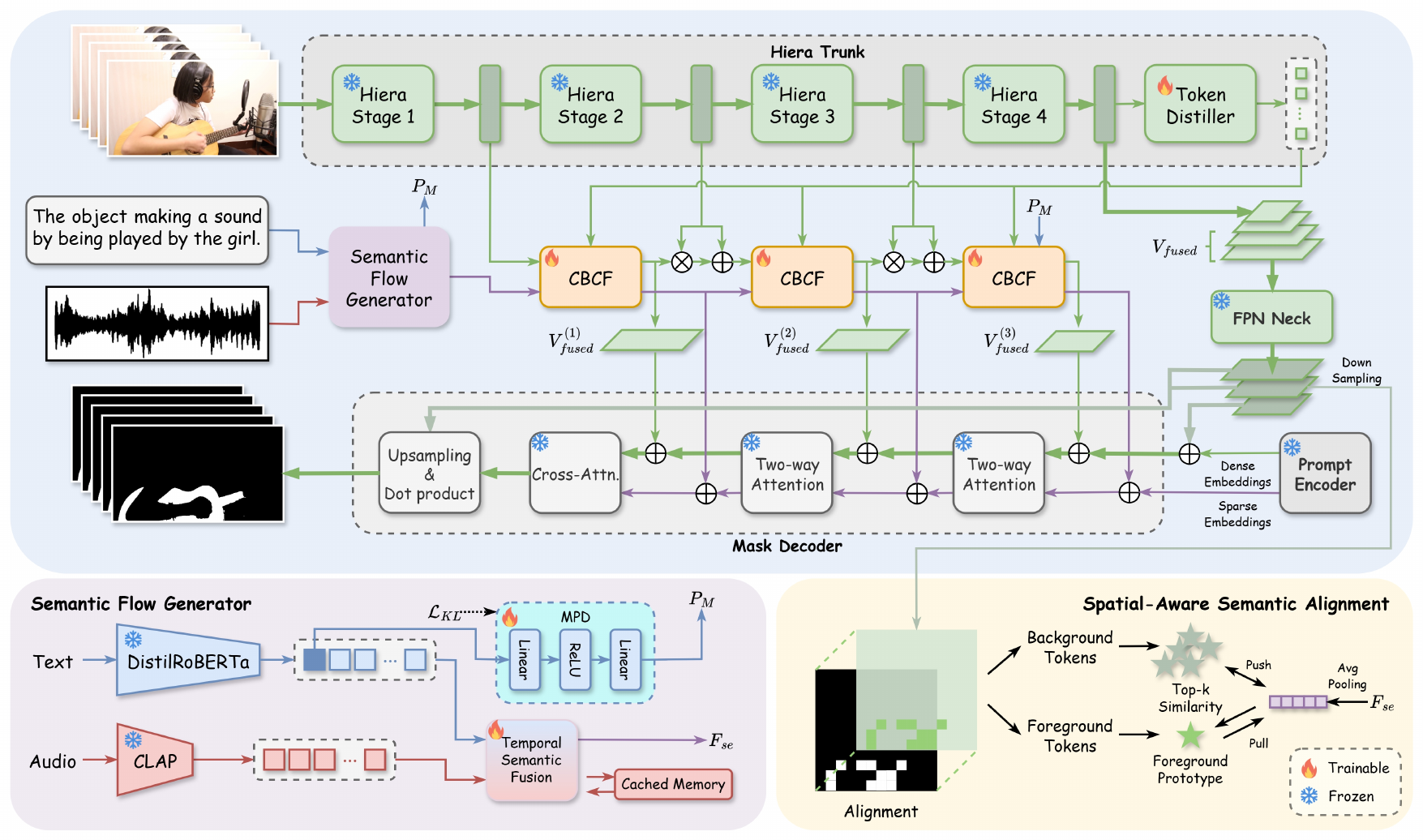}
  \caption{Overview of PRIMED. Multimodal features extracted from encoders, together with distilled visual tokens, are fused through stage-wise CBCF modules to generate embeddings for the mask decoder. The modality prior from MPD modulates the highest-level CBCF module. During training, the SASA loss further enforces semantic discrimination between foreground and background regions.}
  \label{fig:framework}
\end{figure*}

\subsection{Referring Audio-Visual Segmentation}
Referring Audio-Visual Segmentation (Ref-AVS) extends AVS by introducing natural language expressions as additional conditioning signals, requiring models to localize targets under visual, auditory, and linguistic cues.

Early approaches, such as EEMC~\cite{wang2024ref} and VoCa~\cite{guo2025leveraging}, focus on learning unified multimodal representations, where cross-modal features are jointly encoded for segmentation, with VoCa further enhancing spatio-temporal modeling via Mamba-based architectures. With the emergence of visual foundation models, a line of work leverages prompt-based paradigms to adapt SAM/SAM2 for Ref-AVS. TSAM~\cite{radman2025tsam} incorporates temporal modeling and multimodal prompts, while SAM2-LOVE~\cite{wang2025sam2} compresses multimodal cues into a unified prompt token to guide mask decoding.

Beyond these two lines, recent efforts have begun to improve higher-level semantic understanding. Crab~\cite{du2025crab} extends the problem toward unified audio-visual scene understanding with explicit cooperation, whereas AURORA~\cite{luo2025aurora} and TGS-Agent~\cite{zhou2025think} incorporate structured reasoning or staged processing to better handle complex scenarios. In parallel, omnimodal frameworks such as Omni-R1~\cite{zhong2025omni} and OISA~\cite{ying2025towards} further broaden the multimodal reasoning setting by incorporating multimodal large language models.

Despite these advances, existing methods predominantly treat multimodal inputs as signals to be fused, prompted, or reasoned over, while overlooking the varying importance of different modalities in guiding target selection. As a result, they struggle to resolve ambiguities when multiple candidate regions partially satisfy the referring expression.

\textbf{In summary}, existing Ref-AVS methods largely treat all modalities as homogeneous cues and fail to explicitly account for their varying relevance in target selection. Consequently, they often struggle when multiple candidate regions partially satisfy the referring expression, leading to ambiguous or incorrect segmentation. This limitation motivates revisiting Ref-AVS from the perspective of \textbf{biased competition}, emphasizing the need to leverage the modality-aware prior and hierarchical visual context to progressively resolve ambiguity and guide target selection. This perspective directly informs the design of our proposed PRIMED framework.

\section{Method}
PRIMED is a Ref-AVS framework built on a frozen SAM2. As shown in Fig.~\ref{fig:framework}, text and audio features are first aligned into a unified semantic flow, together with a language-guided modality prior predicted by the \textbf{Modality Prior Decoder (MPD)}. The semantic flow then interacts with the multi-scale visual hierarchy of the Hiera trunk through stage-wise \textbf{Cross-modal Biased Competition Fusion (CBCF)} modules, where the modality prior modulates the highest-level competition and distilled visual tokens provide shared global context. Finally, the fused visual features are fed into the FPN-style neck and mask decoder for segmentation, while a \textbf{Spatial-Aware Semantic Alignment (SASA)} loss is introduced during training to enhance foreground-background discrimination.

\subsection{Semantic Flow Generator}
Given the video frames, the referring expression, and the corresponding audio waveform, we first extract modality-specific representations using frozen pretrained encoders. Formally, let $\mathcal{V} = \{v_i\}_{i=1}^{T}$ denote the input video consisting of $T$ consecutive frames sampled at 1-second intervals, where each frame $v_i \in \mathbb{R}^{3 \times H \times W}$ has a resolution of ${H} \times {W}$. The video frames $\mathcal{V}$ are then fed into the Hiera trunk~\cite{ryali2023hiera} to extract hierarchical visual features at four stages, denoted as $\{F_V^{(n)}\}_{n=1}^{4}$, where $F_V^{(n)} \in \mathbb{R}^{T \times C^{(n)} \times H^{(n)} \times W^{(n)}}$, with $C^{(n)}$, $H^{(n)}$, and $W^{(n)}$ representing the channel dimension, height, and width of the visual feature map at the $n$-th stage respectively. The corresponding audio waveform, with a duration of $T$ seconds, is encoded by a pretrained CLAP~\cite{elizalde2023clap}. Specifically, the waveform is first converted into a log-Mel spectrogram with 64-bin filter banks, and then processed by the HTS-AT~\cite{chen2022hts} to obtain audio embeddings $F_A \in \mathbb{R}^{T \times d_A}$, where $d_A$ denotes the audio feature dimension. The referring expression is processed by a pretrained DistilRoBERTa~\cite{liu2019roberta,sanh2019distilbert}, producing text embeddings $F_T \in \mathbb{R}^{L \times d_T}$, where $L$ is the number of tokens in the expression and $d_T$ is the text feature dimension.

\begin{figure}[t]
  \centering
  \includegraphics[width=0.45\textwidth]{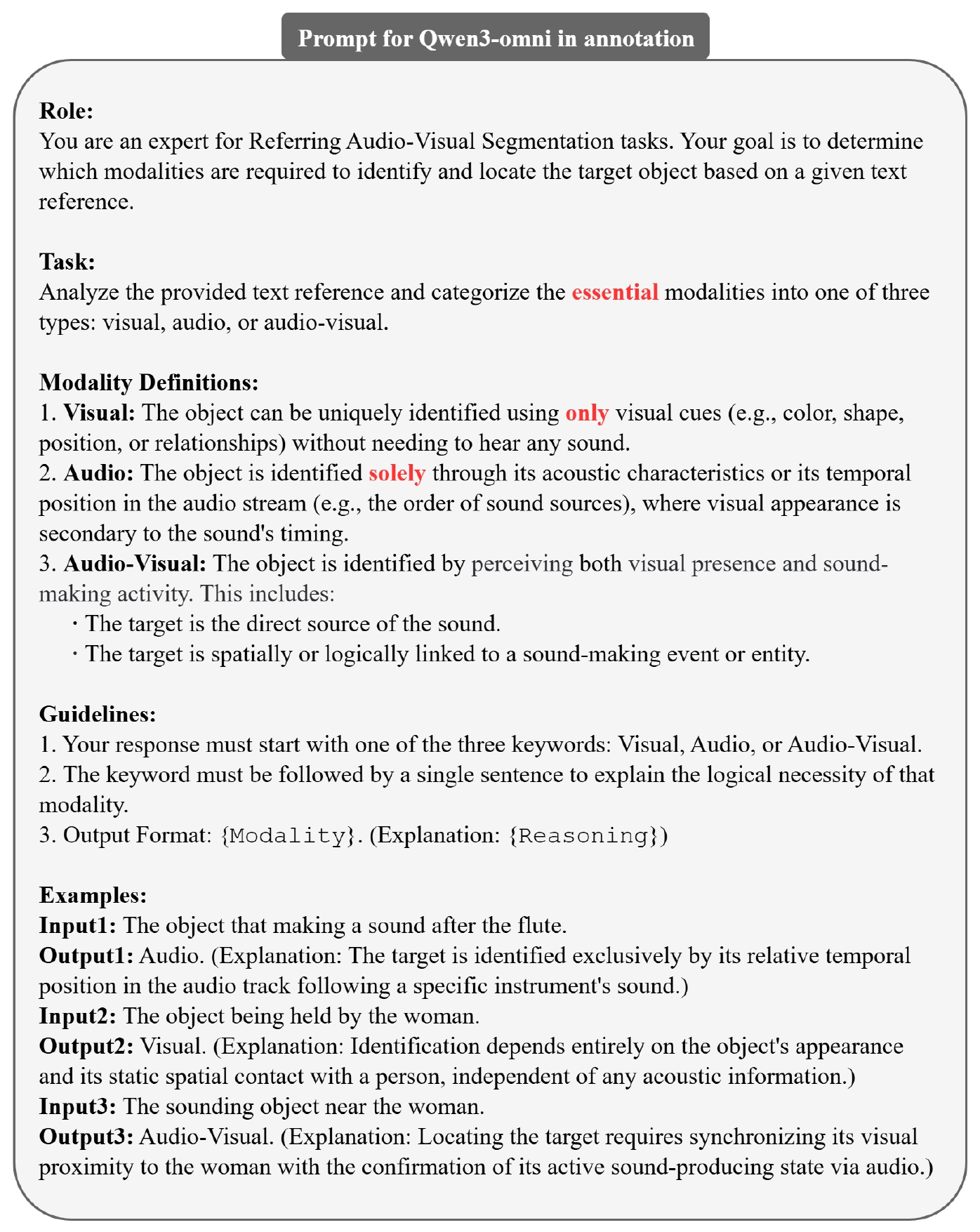}
  \caption{Prompts used for the label annotation of Modality Prior Decoder.}
  \label{fig:annotation}
\end{figure}

\begin{figure}[t]
  \centering
  \includegraphics[width=0.45\textwidth]{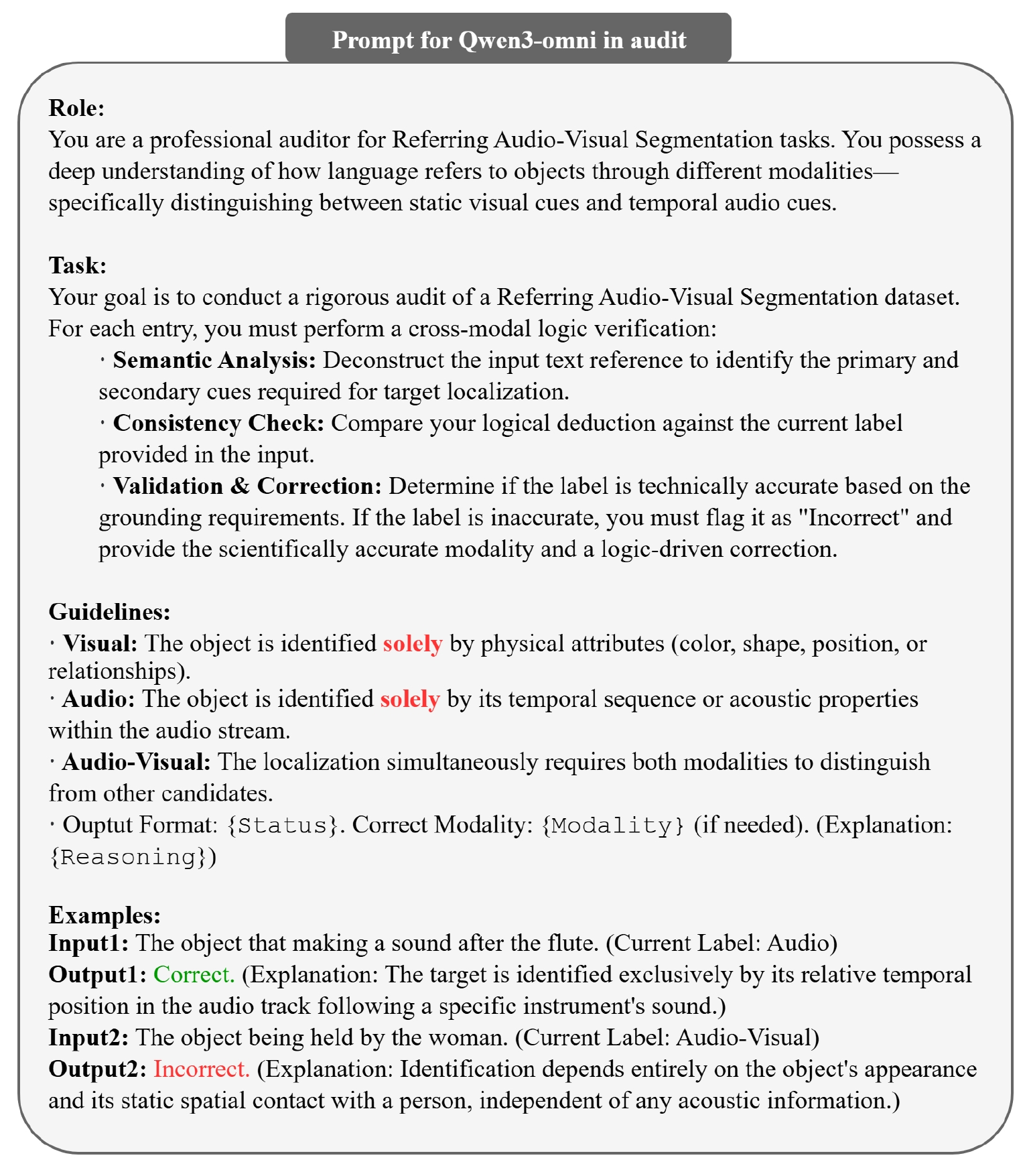}
  \caption{Prompts used for the label annotation audit.}
  \label{fig:audit}
\end{figure}

\textbf{Modality Prior Decoder.}
The Modality Prior Decoder (MPD) is designed to predict a modality prior $P_M$ that characterizes the relative importance of visual and auditory cues for the given referring expression. We argue that direct fuzzy matching between words and modalities is often unreliable and affects the generalization performance. Therefore, the required modality cannot be robustly determined from surface words alone, but should instead be inferred from the sentence semantics in a context-aware manner. Specifically, we take the first token of $F_T$ as the global textual representation $t_g \in \mathbb{R}^{d_T}$, and feed it into MPD to produce three logits corresponding to audio-dominant, visual-dominant, and audio-visual joint cases. The MPD is a lightweight MLP, which consists of two linear layers with a ReLU activation in between. The whole process can be formulated as follows:
\begin{equation}
z_M = W_2 \, \sigma(W_1 t_g + b_1) + b_2,
\end{equation}
\begin{equation}
P_M = \mathrm{Softmax}(z_M) = [p_A, p_V, p_{AV}],
\end{equation}
where $\sigma$ is ReLU, $p_A$, $p_V$, and $p_{AV}$ denote the probabilities of three cases above respectively, satisfying
$p_A + p_V + p_{AV} = 1$.

Since Ref-AVS does not provide explicit annotations on modality relevance, we use Qwen3-omni~\cite{xu2025qwen3} to generate soft labels for MPD and further perform an additional audit to improve reliability. The prompt provided in Fig.~\ref{fig:annotation} is used to annotate the labels for MPD. The audit prompt, shown in Fig.~\ref{fig:audit}, further verifies the annotated labels through an additional round of consistency checking, improving the reliability of the supervision. If the audit result differs from the initial annotation, the referring expression is finally examined by a human annotator. Compared with hard one-hot labels, such supervision better captures gradual modality preference and ambiguous cases. For each referring expression, Qwen3-omni estimates whether the target can be primarily localized from audio, vision, or from their joint reasoning:
\begin{equation}
\widetilde{P}_M = [\tilde{p}_A, \tilde{p}_V, \tilde{p}_{AV}].
\end{equation}

We optimize the MPD with a Kullback-Leibler divergence:
\begin{equation}
\mathcal{L}_{KL} = \sum_{c \in \{A,V,AV\}} \widetilde{P}_M^{(c)} \log \frac{\widetilde{P}_M^{(c)}}{P_M^{(c)}}.
\end{equation}

\textbf{Temporal Semantic Fusion.}
After obtaining the modality prior, we fuse text and audio representations into a unified semantic flow. Both modalities are first projected into a shared latent space, denoted as $\widetilde{F}_A$ and $\widetilde{F}_T$. The features are then concatenated and processed by a Transformer block:
\begin{equation}
S_0 = \mathrm{Concat}(\widetilde{F}_A, \widetilde{F}_T),
\end{equation}
\begin{equation}
\bar{S} = \mathrm{LN}\big(S_0 + \mathrm{MHSA}(S_0)\big),
\end{equation}
\begin{equation}
S = \mathrm{LN}\big(\bar{S} + \mathrm{FFN}(\bar{S})\big),
\end{equation}
where $\mathrm{Concat}(\cdot)$ denotes the concatenation, $\mathrm{LN}(\cdot)$ denotes the LayerNorm, and $\mathrm{MHSA}(\cdot)$ and $\mathrm{FFN}(\cdot)$ denote multi-head self-attention and feed-forward network respectively. We then split $S$ into the refined audio and text representations.

Following~\cite{wang2024ref}, we introduce a cached memory on the audio branch to emphasize the temporal mutations. Different from the recurrent update, our implementation computes cached memories for all frames in parallel over the whole temporal sequence. In this way, the current audio representation is amplified when it deviates significantly from the historical context, making the semantic flow more responsive to audio events. 

In particular, for the $i$-th second, the cached memory stores the mean of preceding audio features:
\begin{equation}
C_A^{(i)} =
\begin{cases}
0, & i = 1,\\
\frac{1}{i-1}\sum_{j=1}^{i-1} S_A^{(j)}, & i > 1.
\end{cases}
\end{equation}

The enhanced audio representation is then computed for all frames simultaneously as follows:
\begin{equation}
\hat{S}_A = (\beta + 1) S_A - \beta C_A,
\end{equation}
where $\beta$ is a hyper-parameter.

Finally, the temporally enhanced audio representation $\hat{S}_A$ and the token-level text representation are concatenated and then broadcast along the temporal dimension to form a unified semantic flow $F_{se} \in \mathbb{R}^{T \times (T + L) \times d}$.

\subsection{Token Distiller}
Under the biased competition view, stable target localization requires not only local stage-wise responses, but also a compact high-level visual bias to guide competition toward globally consistent object semantics. To this end, we introduce a Token Distiller to compress the highest-level Hiera features into a set of compact visual tokens, which serve as global visual biases shared by all CBCF modules. We first flatten and project the final-stage visual feature map into the distilled-token embedding space:
\begin{equation}
\hat{F}_{V}^{(4)} = \psi(F_V^{(4)}) \in \mathbb{R}^{T \times M_4 \times d_0},
\end{equation}
where $\psi(\cdot)$ is a projection layer, $M_4 = H^{(4)} \times W^{(4)}$, and $d_0$ is the token embedding dimension.

We then introduce a set of learnable distilled tokens $V_0 \in \mathbb{R}^{K \times d_0}$, where $K$ denotes the number of distilled tokens. These tokens interact with the projected visual tokens through cross-attention to extract compact global visual cues. We do not apply LayerNorm in this process, since feature normalization may destroy the orthogonality structure among distilled tokens:
\begin{equation}
\hat{V}_{dis} = V_0 + \mathrm{MHCA}(V_0, \hat{F}_{V}^{(4)}, \hat{F}_{V}^{(4)}),
\end{equation}
where $\mathrm{MHCA}(\cdot)$ denotes multi-head cross-attention. 

The distilled tokens are further refined by an MLP with a residual connection:
\begin{equation}
{V}_{dis} = \hat{V}_{dis} + \mathrm{MLP}(\hat{V}_{dis}),
\end{equation}

To encourage diversity and reduce redundancy among different distilled tokens, we impose Gram-Schmidt orthogonalization on ${V}_{dis}$ during training. The resulting distilled tokens are shared across all CBCF modules, allowing local stage-wise features to interact not only with the semantic flow, but also with compact global visual context distilled from the top-most visual representation.

\begin{figure}[t]
  \centering
  \includegraphics[width=\linewidth]{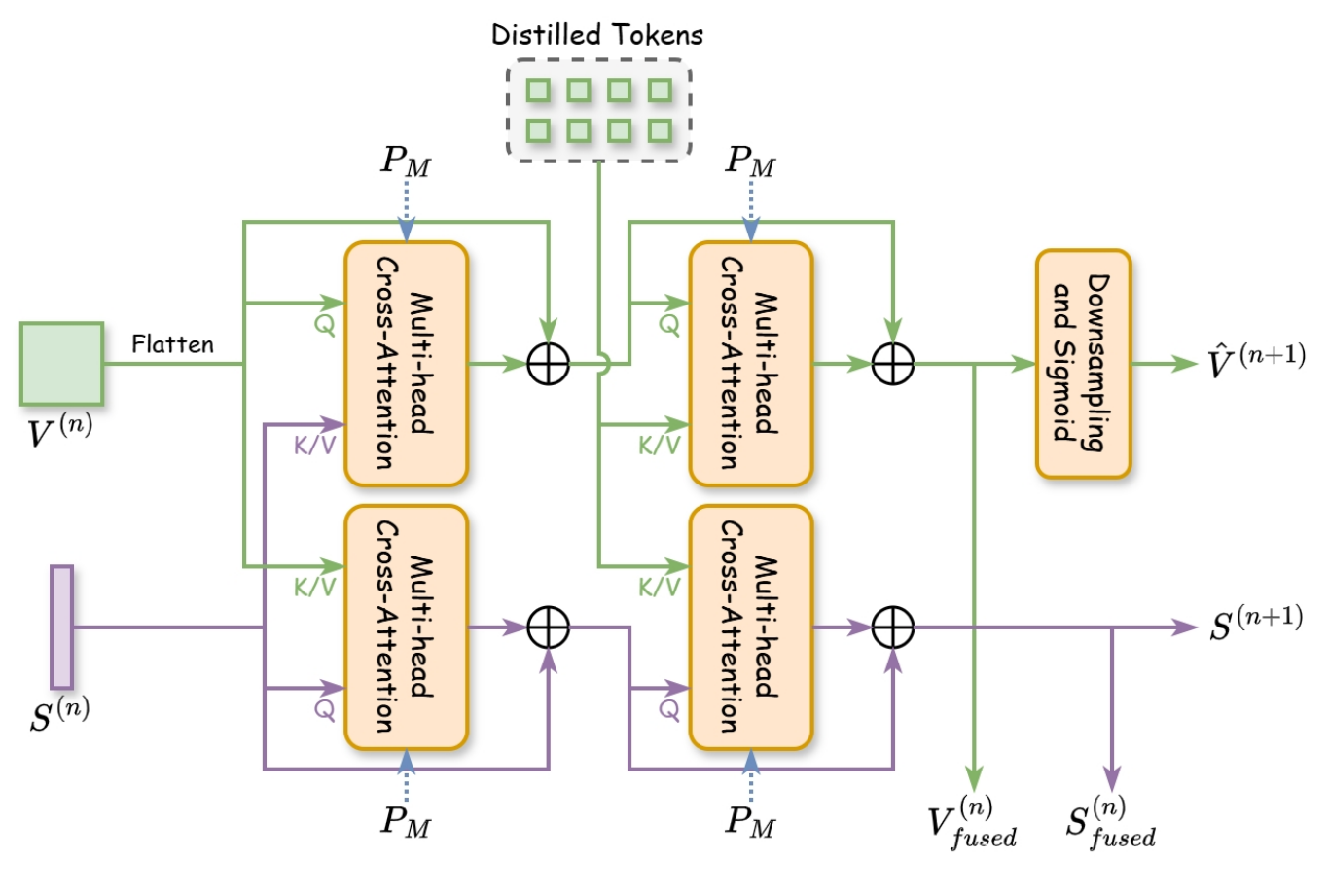}
  \caption{Illustration of the CBCF module at stage 1 and 2. The structure of CBCF is similar at stage 3, but the modality prior $P_M$ is injected and the paths towards the $\hat{V}^{(4)}$ and ${S}^{(4)}$ are removed.}
  \label{fig:CBCF}
\end{figure}

\subsection{Cross-modal Biased Competition Fusion}
Each Cross-modal Biased Competition Fusion (CBCF) module models the competition between stage-wise visual cues and query-conditioned semantic cues under a shared global visual bias, illustrated in Fig.~\ref{fig:CBCF}. For the $n$-th stage, let $V^{(n)}$ and $S^{(n)}$ denote the input visual feature and semantic flow, respectively. After projection into a shared space, CBCF performs two rounds of bidirectional multi-head cross-attention. Specifically, in the second round, both branches further attend to the distilled tokens $V_{dis}$, which provide shared global visual bias and encourage consistent competition across stages.

We reshape the updated visual branch as $V_{fused}^{(n)}$ and directly take the updated semantic branch as $S_{fused}^{(n)}$. For $n < 3$, the fused visual branch is further downsampled and used to modulate the next-stage visual feature, while the semantic branch is directly propagated to the next stage.

Stage 3 differs from previous stages in two aspects. First, the paths toward $\hat{V}^{(4)}$ and $S^{(4)}$ are removed, since this is the final fusion stage before mask decoding. Second, the modality prior $P_M$ is injected into all MHCA blocks as an additive attention bias to regulate the final competition between visual and semantic cues. Specifically, we project the original text embeddings $F_T$, audio embeddings $F_A$, and stage-4 visual features $F_V^{(4)}$ into a shared space. Let $\tilde{F}_V$ denote the projected stage-4 visual feature, $T_g$ denote the projected global text token derived from $t_g$, and $A_g$ denote the global audio feature obtained by average pooling over projected $F_A$. We then compute visual-text and visual-audio similarities:
\begin{equation}
\mathrm{Sim}_{vis}=\frac{\langle T_g,\tilde{F}_V\rangle+1}{2}, \quad
\mathrm{Sim}_{aud}=\frac{\langle A_g,\tilde{F}_V\rangle+1}{2}.
\end{equation}

These two similarities are combined by the modality prior into a unified score
\begin{equation}
\hat{P}=p_V \mathrm{Sim}_{vis} + p_A \mathrm{Sim}_{aud} + p_{AV}(\mathrm{Sim}_{vis}\odot \mathrm{Sim}_{aud}),
\end{equation}
which is further converted into a logit-space bias:
\begin{equation}
b_M=\gamma_p \log \frac{\hat{P}}{1-\hat{P}},
\end{equation}
where $\gamma_p$ is a learnable scaling factor.

The bias $b_M$ is broadcast and added to the attention logits before softmax:
\begin{equation}
\mathrm{MHCA}_{P}(Q,K,V)=\mathrm{Softmax}\!\left(\frac{QK^\top}{\sqrt{d}}+b_M\right)V.
\end{equation}

We apply the modality prior only at the highest CBCF stage because it is a global semantic cue that is better aligned with high-level visual semantics than low-level local patterns.

The outputs of CBCF are further used to prompt the frozen SAM2 decoder, and both dense and sparse features are used to prompt the frozen SAM2 mask decoder. Specifically, the stage-wise fused visual features $\{V_{fused}^{(n)}\}_{n=1}^{3}$ serve as dense visual prompts and are fed into the FPN neck, while the fused semantic representations $\{S_{fused}^{(n)}\}_{n=1}^{3}$ are injected as sparse semantic prompts. 

For the visual branch, each fused visual feature is transformed by a lightweight convolution followed by average pooling to match the shape of the dense prompt embedding:
\begin{equation}
d^{(n)} = \mathrm{AvgPool}(\mathrm{Conv}(V_{fused}^{(n)})).
\end{equation}

For the semantic branch, each fused semantic representation is first projected by a linear layer, then averaged over the token dimension, and finally broadcast to match the shape of the sparse prompt embedding:
\begin{equation}
s^{(n)} = \mathrm{Broadcast}(\mathrm{MeanPool}(\mathrm{Linear}(S_{fused}^{(n)}))).
\end{equation}

Let $\hat{E}_d^{(n)}$ and $\hat{E}_s^{(n)}$ denote the prompt embeddings before injection. The updated dense and sparse prompt embeddings are obtained by element-wise addition:
\begin{equation}
E_d^{(n)}=\hat{E}_d^{(n)}+d^{(n)}, \qquad
E_s^{(n)}=\hat{E}_s^{(n)}+s^{(n)}.
\end{equation}

To better match feature granularity, the three groups of fused representations are sequentially injected into the decoder hierarchy. In particular, higher-level fused representations are injected into earlier decoder blocks, while lower-level fused representations are introduced at later decoding steps closer to final mask prediction.

\subsection{Spatial-Aware Semantic Alignment}
To further encourage the semantic flow to focus on the true target region, we introduce a Spatial-Aware Semantic Alignment (SASA) loss. Specifically, after discarding the lowest-resolution FPN output, we select the second retained FPN feature map and denote it by $F_{fpn} \in \mathbb{R}^{C \times H_f \times W_f}$. We first downsample both $F_{fpn}$ and the corresponding ground truth $Y_{gt} \in \{0,1\}^{H \times W}$ to a fixed spatial size of $64 \times 64$. Next, we flatten the downsampled feature map into a token sequence, the downsampled ground-truth mask $\tilde{Y}_{gt}$ is used to partition the token sequence into foreground and background token sets, denoted as $\Omega_f$ and $\Omega_b$:
\begin{equation}
\mathcal{I}_f = \{i \mid \tilde{Y}_i = 1\}, \qquad
\mathcal{I}_b = \{i \mid \tilde{Y}_i = 0\}.
\end{equation}
\begin{equation}
\Omega_f = \{z_i \mid i \in \mathcal{I}_f\}, \qquad
\Omega_b = \{z_i \mid i \in \mathcal{I}_b\}.
\end{equation}

For each frame, we further aggregate the semantic flow $F_{se}$ generated by the Semantic Flow Generator by average pooling over all its tokens, so as to obtain one semantic anchor token per frame, denoted as $\hat{F}_{se}$. Given the foreground token set $\Omega_f$ and background token set $\Omega_b$, we first construct the foreground prototype by average pooling over all foreground tokens:
\begin{equation}
p_f = \frac{1}{|\Omega_f|} \sum_{z_i \in \Omega_f} z_i.
\end{equation}

For the background region, after the projection and normalization, we compute the cosine similarity between $\hat{F}_{se}$ and each background token, and choose the top-$k$ most similar background tokens:
\begin{equation}
\Omega_{b,k} = \mathrm{TopK}_{z_i \in \Omega_b} \big( \mathrm{sim}(\hat{F}_{se}, z_i) \big),
\end{equation}

Finally, we treat the foreground prototype $p_f$ as the positive sample and the selected hard background tokens $\Omega_{b,k}$ as negatives. The SASA loss is defined as:
\begin{equation}
\mathcal{L}_{\mathrm{SASA}}
=
-\log
\frac{
\phi(\hat{F}_{se}, p_f)
}{
\phi(\hat{F}_{se}, p_f)
+
\sum_{z_i \in \Omega_{b,k}}
\phi(\hat{F}_{se}, z_i)
},
\end{equation}
where $\phi(a,b) = \exp(\mathrm{sim}(a,b)/\tau)$, and $\tau$ is the temperature parameter.

Therefore, the semantic anchor of each frame is encouraged to align with the coarse foreground prototype while being pushed away from the most semantically similar background tokens. Such a spatially aware contrastive objective explicitly strengthens the consistency between the semantic flow and the target-related visual region, and suppresses confusing responses from background areas.

\subsection{Training Objectives}
The overall loss function is defined as
\begin{equation}
\mathcal{L}
=
\mathcal{L}_{seg}
+
\lambda_{SASA}\mathcal{L}_{SASA}
+
\lambda_{KL}\mathcal{L}_{KL}
+
\lambda_{orth}\mathcal{L}_{orth},
\end{equation}
where $\lambda_{SASA}$, $\lambda_{KL}$, $\lambda_{orth}$ are balancing coefficients, and $\mathcal{L}_{seg}$ denotes the segmentation loss. In our implementation, $\mathcal{L}_{seg}$ is defined as the sum of binary cross-entropy loss and Dice loss:
\begin{equation}
\mathcal{L}_{seg} = \mathcal{L}_{BCE} + \mathcal{L}_{Dice}.
\end{equation}

\section{Experiments}
\subsection{Experimental Settings}
\noindent
\textbf{Dataset.}
We evaluate our method on the Ref-AVS benchmark~\cite{wang2024ref}, which contains 4,000 video clips with a total duration of 10 seconds and 20,000 referring expressions. The dataset is divided into three sets: 2,908 videos for training, 276 videos for validation, and 818 videos for testing. The test set is further divided into three subsets: \textit{seen} split, containing 292 cases with the same object classes as the training set; \textit{unseen} split, containing 269 cases with additional 13 classes not present in the training set; \textit{null} split, containing 257 cases where the referred target does not exist in the video. In \textit{null} split, the model is required to generate empty masks.

\noindent
\textbf{Evaluation Metrics.}
Following~\cite{wang2024ref}, we adopt the Jaccard Index ($\mathcal{J}$), F-score ($\mathcal{F}$), and their average ($\mathcal{J}\&\mathcal{F}$) as primary evaluation metrics. For the null split, we employ the metric $\mathcal{S}$, which is the square root of the ratio between the number of predicted foreground pixels and the background pixels. 

\noindent
\textbf{Implementation Details.}
Our framework is built on the frozen Hiera-Large variant of SAM2, with all input frames resized to $1024\times1024$. During training, prompts are provided for all frames to enforce frame-level supervision, while at inference only the first frame is prompted and subsequent frames are segmented using the original SAM2 memory-based propagation.

For text encoding, we use DistilRoBERTa as the text encoder. Specifically, we adopt the \textit{distilroberta-base} checkpoint from HuggingFace. DistilRoBERTa is a distilled variant of RoBERTa-base that provides a lightweight yet effective language encoder. In our framework, it outputs token-level textual embeddings with the hidden dimension of 768, from which the first token is further used as the global textual representation.

For audio encoding, we follow the official CLAP configuration. Since each video in the dataset is 10 seconds long, we set $T = 10$ by sampling one frame per second. Specifically, the input audio is converted into a 64-bin log-Mel spectrogram with a sampling rate of 44.1 kHz, a hop size of 320, a window size of 1024, and a frequency range of 50--8000 Hz, and is then processed by the HTS-AT audio encoder. The resulting audio embedding dimension is 1024.

We use 4 distilled visual tokens with dimension 512, set $k = 10$ for background token selection, and train the model on a single RTX 5880Ada GPU with AdamW at a base learning rate of $5\times10^{-5}$. The first 5\% of training iterations use linear warmup with a start factor of 0.1, followed by cosine annealing. The model is trained for 15 epochs with batch size 1, and the loss weights are set to $\lambda_{SASA}:\lambda_{KL}:\lambda_{orth}=5:1:1$. The $\beta$ in the cached memory is fixed to 1 in all experiments.

\begin{table*}[!t]
  \centering
  \caption{Performance comparison with state-of-the-art methods on the Ref-AVS benchmark. $^{*}$ denotes methods adapted to the tri-modal setting. The mix means the average value of seen and unseen splits.}
  \label{tab:main}
  \begin{tabular*}{\textwidth}{@{\extracolsep{\fill}}lccccccccccc}
  \toprule
  \multirow{2}{*}{\textbf{Model}} & \multirow{2}{*}{\textbf{Task}} & \multicolumn{3}{c}{\textbf{Seen (S)}$\,\uparrow$} & \multicolumn{3}{c}{\textbf{Unseen (U)}$\,\uparrow$} & \multicolumn{3}{c}{\textbf{Mix (S+U)}$\,\uparrow$} & \textbf{Null}$\,\downarrow$ \\
   &  & $\mathcal{J}$ & $\mathcal{F}$ & $\mathcal{J}\&\mathcal{F}$ & $\mathcal{J}$ & $\mathcal{F}$ & $\mathcal{J}\&\mathcal{F}$ & $\mathcal{J}$ & $\mathcal{F}$ & $\mathcal{J}\&\mathcal{F}$ & $\mathcal{S}$ \\
  \midrule
  AVSBench~\cite{zhou2022audio} & AVS$^{*}$ & 23.2 & 51.1 & 37.2 & 32.4 & 54.7 & 43.5 & 27.8 & 52.9 & 40.3 & 0.208 \\
  AVSegFormer~\cite{gao2024avsegformer} & AVS$^{*}$ & 33.5 & 47.0 & 40.2 & 36.1 & 50.1 & 43.1 & 34.8 & 48.6 & 41.7 & 0.171 \\
  GAVS~\cite{wang2024prompting} & AVS$^{*}$ & 28.9 & 49.8 & 39.4 & 29.8 & 49.7 & 39.8 & 29.4 & 49.8 & 39.6 & 0.190 \\
  \midrule
  ReferFormer~\cite{wu2022language} & Ref-VOS$^{*}$ & 31.3 & 50.1 & 40.7 & 30.4 & 48.8 & 39.6 & 30.9 & 49.5 & 40.2 & 0.176 \\
  R2VOS~\cite{li2023robust} & Ref-VOS$^{*}$ & 25.0 & 41.0 & 33.0 & 27.9 & 49.8 & 38.9 & 26.5 & 45.4 & 35.9 & 0.183 \\
  \midrule
  EEMC~\cite{wang2024ref} & Ref-AVS & 34.2 & 51.3 & 42.8 & 49.5 & 64.8 & 57.2 & 41.9 & 58.1 & 50.0 & \underline{0.007} \\
  VoCa~\cite{guo2025leveraging} & Ref-AVS & 40.9 & 60.8 & 50.9 & 51.7 & 70.8 & 61.2 & 46.3 & 66.8 & 56.6 & \textbf{0.005} \\
  SAM2-LOVE~\cite{wang2025sam2} & Ref-AVS & 43.5 & 51.9 & 47.7 & 66.5 & 72.3 & 69.4 & 55.0 & 62.1 & 58.5 & 0.23 \\
  TSAM~\cite{radman2025tsam} & Ref-AVS & 43.4 & 56.8 & 50.1 & 54.6 & 66.4 & 60.5 & 49.0 & 61.6 & 55.3 & 0.017 \\
  Crab~\cite{du2025crab} & Ref-AVS & 40.5 & 58.0 & 49.3 & 45.6 & 63.0 & 54.3 & 43.1 & 60.5 & 46.2 & - \\
  TGS-Agent~\cite{zhou2025think} & Ref-AVS & 49.5 & 60.4 & 54.9 & \textbf{73.2} & \textbf{80.6} & \textbf{76.9} & 61.3 & 70.5 & 65.9 & 0.035 \\
  OISA-1B~\cite{ying2025towards} & Omni-AVS & 51.7 & 58.7 & 55.2 & 58.3 & 65.1 & 61.7 & 54.5 & 61.4 & 58.0 & 0.098 \\
  Omni-R1~\cite{zhong2025omni} & Omni-Reasoning & 43.0 & 51.4 & 47.2 & 71.3 & \underline{77.0} & \underline{74.2} & 57.2 & 64.2 & 60.7 & - \\
  AURORA~\cite{luo2025aurora} & Ref-AVS & \underline{63.2} & \textbf{72.8} & \underline{68.0} & 69.7 & 76.4 & 73.0 & \underline{66.5} & \textbf{74.6} & \underline{70.1} & - \\
  \midrule
  \textbf{PRIMED (Ours)} & \textbf{Ours} & \textbf{66.0} & \underline{71.5} & \textbf{68.8} & \underline{71.8} & 74.3 & 73.1 & \textbf{68.9} & \underline{72.9} & \textbf{70.9} & 0.015 \\
  \bottomrule
  \end{tabular*}
\end{table*}

\subsection{Main Results}
We compare our method with prior AVS and Ref-VOS baselines incorporating text and audio modalities, dedicated Ref-AVS approaches, and recent omnimodal frameworks. As shown in Table~\ref{tab:main}, PRIMED achieves the best overall mixed $\mathcal{J}$ and $\mathcal{J}\&\mathcal{F}$, while also obtaining the best $\mathcal{J}$ and $\mathcal{J}\&\mathcal{F}$ performance on the seen split. Compared with AURORA~\cite{luo2025aurora}, PRIMED improves $\mathcal{J}$ by 2.8\% on the seen split and 2.1\% on the unseen split. Our method also shows clear and consistent improvements over other SAM-based Ref-AVS methods. Relative to SAM2-LOVE~\cite{wang2025sam2} and TSAM~\cite{radman2025tsam}, PRIMED achieves clear gains on the seen split, while on the unseen split, it improves $\mathcal{J}$ and $\mathcal{F}$ by 5.3\% and 2.0\% over SAM2-LOVE, and by 17.2\% and 7.9\% over TSAM respectively. These results indicate that explicitly modeling modality-aware competition provides stronger generalization under challenging scenarios, while maintaining robustness in in-domain settings. On the null split, our method slightly falls behind VoCa~\cite{guo2025leveraging} by 0.01 but still outperforms most prior methods, indicating strong suppression when the referred target is absent. Nevertheless, compared with recent MLLM-based reasoning methods such as TGS-Agent~\cite{zhou2025think}, PRIMED still exhibits a gap in F-score and remains a direction for future improvement.

\begin{table}[!t]
  \centering
  \caption{Ablation study on modality prior and distilled visual tokens.}
  \label{tab:ablation_1}
  \begin{tabular*}{\columnwidth}{@{\extracolsep{\fill}}lccccc}
  \toprule
  \multirow{2}{*}{\textbf{Method}} & \multicolumn{2}{c}{\textbf{Seen}} & \multicolumn{2}{c}{\textbf{Unseen}} & \textbf{Null} \\
  & $\mathcal{J}$ & $\mathcal{F}$ & $\mathcal{J}$ & $\mathcal{F}$ & $\mathcal{S}$ \\
  \midrule
  Full & \textbf{66.0} & \textbf{71.5} & \textbf{71.8} & \textbf{74.3} & \textbf{0.015} \\
  w/o prior & 64.4 & 69.7 & 71.1 & 74.2 & 0.015 \\
  w/o prior + distillation & 63.9 & 69.0 & 70.3 & 73.3 & 0.017 \\
  \bottomrule
  \end{tabular*}
\end{table}

\begin{figure}[!t]
  \centering
  \includegraphics[width=\linewidth]{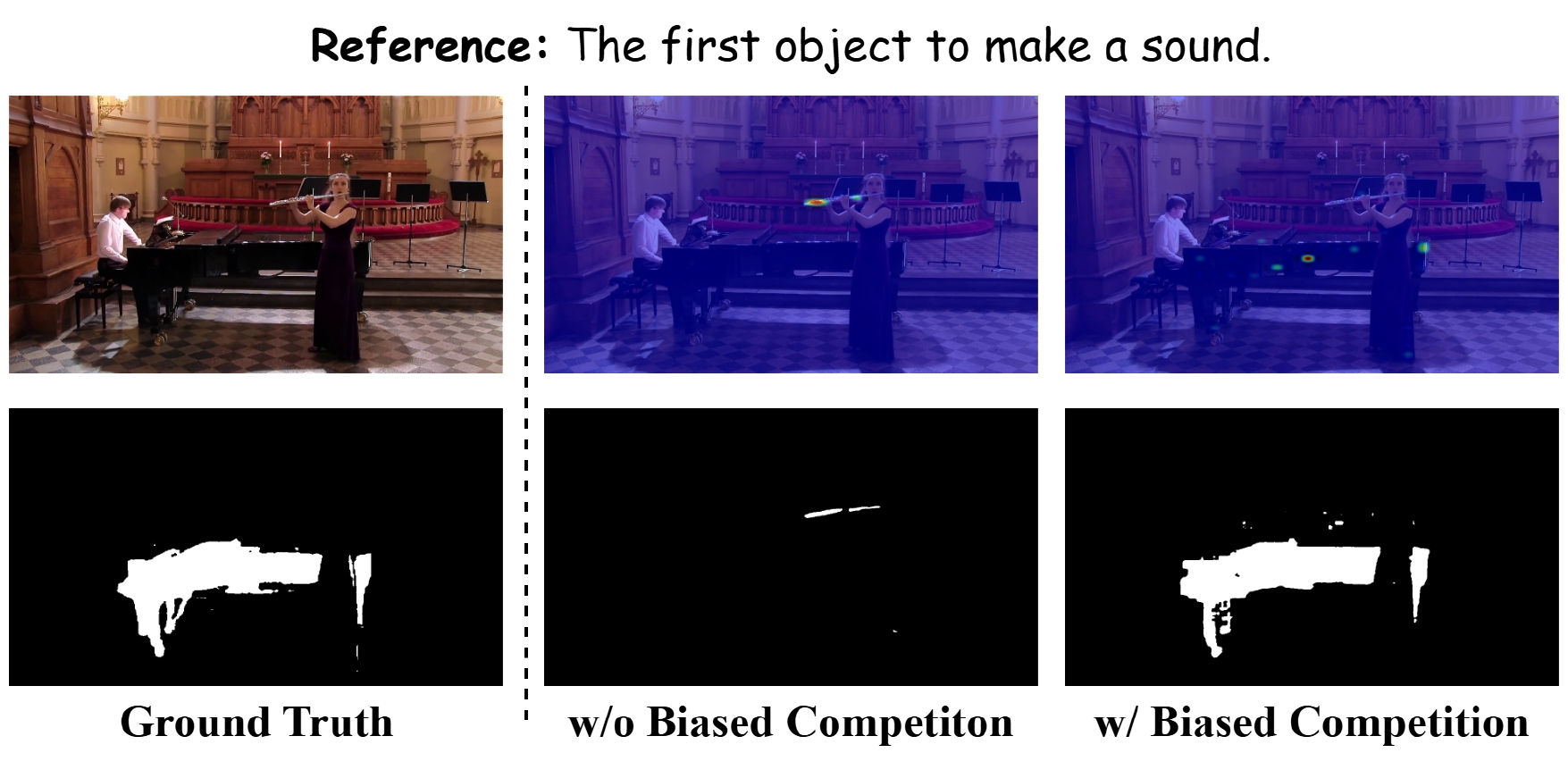}
  \caption{Attention visualization for the biased competition ablation. Without biased competition, attention is more easily drawn to the flute as a competing distractor.}
  \label{fig:ablation}
\end{figure}

\subsection{Ablation Study}
\textbf{Effectiveness of biased competition.}
We first study the contribution of the modality prior and distilled visual tokens. As shown in Table~\ref{tab:ablation_1}, removing the modality prior consistently degrades performance on both seen and unseen splits, with drops of 1.6\% in $\mathcal{J}$ and 1.8\% in $\mathcal{F}$ on the seen split and modest drops on the unseen split. When the whole biased competition design is removed, the degradation becomes larger on both splits, while the null score also worsens by 0.002. This trend is further supported by the attention visualization in Fig.~\ref{fig:ablation}. In this example, the flute acts as a strong distractor because its sound is more salient than that of the piano, while the piano itself is visually less distinctive as it partially blends with the black dress of the woman in front. Without biased competition, the model is easily attracted to this more salient but irrelevant competitor and fails to recover the target region. In contrast, with biased competition, PRIMED better suppresses the distracting flute and assigns more attention to the piano.

\textbf{Effectiveness of sparse and dense embeddings.}
Table~\ref{tab:ablation_2} evaluates the effect of sparse and dense prompt embeddings. When only dense embeddings are retained, performance drops slightly on both seen and unseen splits. However, when only sparse embeddings are used, the degradation becomes much larger on the seen split, reaching 3.3\% in $\mathcal{J}$ and 2.0\% in $\mathcal{F}$, and still leads to drops of 1.5\% in $\mathcal{J}$ and 0.8\% in $\mathcal{F}$ on the unseen split. These results suggest that dense visual prompting plays a more direct role in stable mask prediction, while sparse semantic prompting mainly provides complementary high-level guidance. The best performance is achieved when both prompts are utilized, indicating that accurate segmentation requires both target-aware semantic cues and spatially grounded visual cues.

\begin{table}[!t]
  \centering
  \caption{Ablation study on sparse and dense embeddings.}
  \label{tab:ablation_2}
  \begin{tabular*}{\columnwidth}{@{\extracolsep{\fill}}ccccccc}
  \toprule
  \multirow{2}{*}{\shortstack{\textbf{Sparse}\\\textbf{Embedding}}} & \multirow{2}{*}{\shortstack{\textbf{Dense}\\\textbf{Embedding}}} & \multicolumn{2}{c}{\textbf{Seen}} & \multicolumn{2}{c}{\textbf{Unseen}} & \textbf{Null} \\
  & & $\mathcal{J}$ & $\mathcal{F}$ & $\mathcal{J}$ & $\mathcal{F}$ & $\mathcal{S}$ \\
  \midrule
  \checkmark & \checkmark & \textbf{66.0} & \textbf{71.5} & \textbf{71.8} & \textbf{74.3} & \textbf{0.015} \\
  & \checkmark & 65.1 & 70.7 & 70.8 & 73.7 & 0.019 \\
  \checkmark &  & 62.7 & 69.5 & 70.3 & 73.5 & 0.022 \\
  \bottomrule
  \end{tabular*}
\end{table}

\begin{table}[!t]
  \centering
  \caption{Ablation study on the number of distilled visual tokens.}
  \label{tab:ablation_3}
  \begin{tabular*}{\columnwidth}{@{\extracolsep{\fill}}cccccc}
  \toprule
  \multirow{2}{*}{\shortstack{\textbf{Number of}\\\textbf{distilled tokens}}} & \multicolumn{2}{c}{\textbf{Seen}} & \multicolumn{2}{c}{\textbf{Unseen}} & \textbf{Null} \\
  & $\mathcal{J}$ & $\mathcal{F}$ & $\mathcal{J}$ & $\mathcal{F}$ & $\mathcal{S}$ \\
  \midrule
  8 & 65.2 & 70.6 & \textbf{72.1} & \textbf{75.0} & \textbf{0.014} \\
  4 & \textbf{66.0} & \textbf{71.5} & 71.8 & 74.3 & 0.015 \\
  2 & 64.3 & 70.0 & 70.7 & 74.1 & 0.015 \\
  \bottomrule
  \end{tabular*}
\end{table}

\begin{table}[!t]
  \centering
  \caption{Ablation study on SASA loss and orthogonality regularization.}
  \label{tab:ablation_4}
  \begin{tabular*}{\columnwidth}{@{\extracolsep{\fill}}lccccc}
  \toprule
  \multirow{2}{*}{\textbf{Method}} & \multicolumn{2}{c}{\textbf{Seen}} & \multicolumn{2}{c}{\textbf{Unseen}} & \textbf{Null} \\
  & $\mathcal{J}$ & $\mathcal{F}$ & $\mathcal{J}$ & $\mathcal{F}$ & $\mathcal{S}$ \\
  \midrule
  Full & \textbf{66.0} & \textbf{71.5} & \textbf{71.8} & \textbf{74.3} & \textbf{0.015} \\
  w/o $\mathcal{L}_{SASA}$ & 65.2 & 70.5 & 70.7 & 73.8 & 0.016 \\
  w/o $\mathcal{L}_{SASA} + \mathcal{L}_{orth}$ & 64.8 & 70.0 & 70.4 & 73.7 & 0.016 \\
  \bottomrule
  \end{tabular*}
\end{table}

\textbf{Effectiveness of the number of distilled tokens.}
The number of distilled visual tokens also has a clear impact on performance, shown in Table~\ref{tab:ablation_3}. Using 4 tokens provides the best overall balance. When the number is reduced to 2, the performance drops noticeably on both seen and unseen splits, with a larger decline on the seen split. This suggests that too few tokens create a bottleneck in global context summarization, making the propagated bias insufficient to support stable competition across stages. In contrast, increasing the number to 8 slightly improves the unseen split, indicating enhanced generalization ability, but leads to clear degradation on the seen split. This trend implies that more tokens can enrich global bias signals, yet may also introduce redundancy and weaken the compact guidance required for in-domain discrimination. 

\begin{table}[t]
  \centering
  \caption{Ablation study on cached memory.}
  \label{tab:ablation_5}
  \begin{tabular*}{\columnwidth}{@{\extracolsep{\fill}}lccccc}
  \toprule
  \multirow{2}{*}{\textbf{Method}} & \multicolumn{2}{c}{\textbf{Seen}} & \multicolumn{2}{c}{\textbf{Unseen}} & \textbf{Null} \\
  & $\mathcal{J}$ & $\mathcal{F}$ & $\mathcal{J}$ & $\mathcal{F}$ & $\mathcal{S}$ \\
  \midrule
  w/ cached memory & \textbf{66.0} & \textbf{71.5} & \textbf{71.8} & \textbf{74.3} & \textbf{0.015} \\
  w/o cached memory & 65.9 & 71.2 & 70.6 & 73.7 & 0.023 \\
  \bottomrule
  \end{tabular*}
\end{table}

\textbf{Effectiveness of the loss design.}
The loss ablation in Table~\ref{tab:ablation_4} verifies the importance of the SASA loss and orthogonality regularization. Removing SASA loss decreases the performance by 0.8\% in $\mathcal{J}$ and 1.0\% in $\mathcal{F}$ on the seen split, and by 1.1\% in $\mathcal{J}$ and 0.5\% in $\mathcal{F}$ on the unseen split, indicating that explicit foreground-background separation is beneficial for robust target discrimination. When both SASA loss and orthogonality regularization are removed, the performance drops further. This comparison suggests that orthogonality regularization contributes beyond SASA by encouraging different distilled tokens to encode complementary global bias cues rather than redundant responses. Overall, SASA mainly strengthens semantic discrimination between the target and background, while orthogonality regularization improves the diversity and effectiveness of the shared global biases.

\textbf{Effectiveness of cached memory.}
We remove the entire cached memory branch to examine its effect. As shown in Table~\ref{tab:ablation_5}, its removal causes a pronounced degradation of 0.008 on the null split, while also introducing performance drops on other splits. We argue that many null-split samples contain silent audio, under which frame-wise audio features often remain nearly unchanged and provide limited discriminative evidence. Cached memory alleviates this issue by aggregating more stable global semantic context, which helps suppress spurious responses on irrelevant regions. For seen and unseen splits, its contribution is still beneficial, as the accumulated temporal context improves grounding stability under multimodal ambiguity, with a more noticeable effect on the unseen split.

% \begin{figure*}[!t]
%   \centering
%   \includegraphics[width=\linewidth]{visualization.pdf}
%   \caption{The visualization results of referred objects in the Ref-AVS benchmark. Speaker icons indicate the sounding objects. Additional cases are shown in the appendix.}
%   \label{fig:visualization}
% \end{figure*}

\begin{figure*}[!t]
  \centering
  \includegraphics[width=\linewidth]{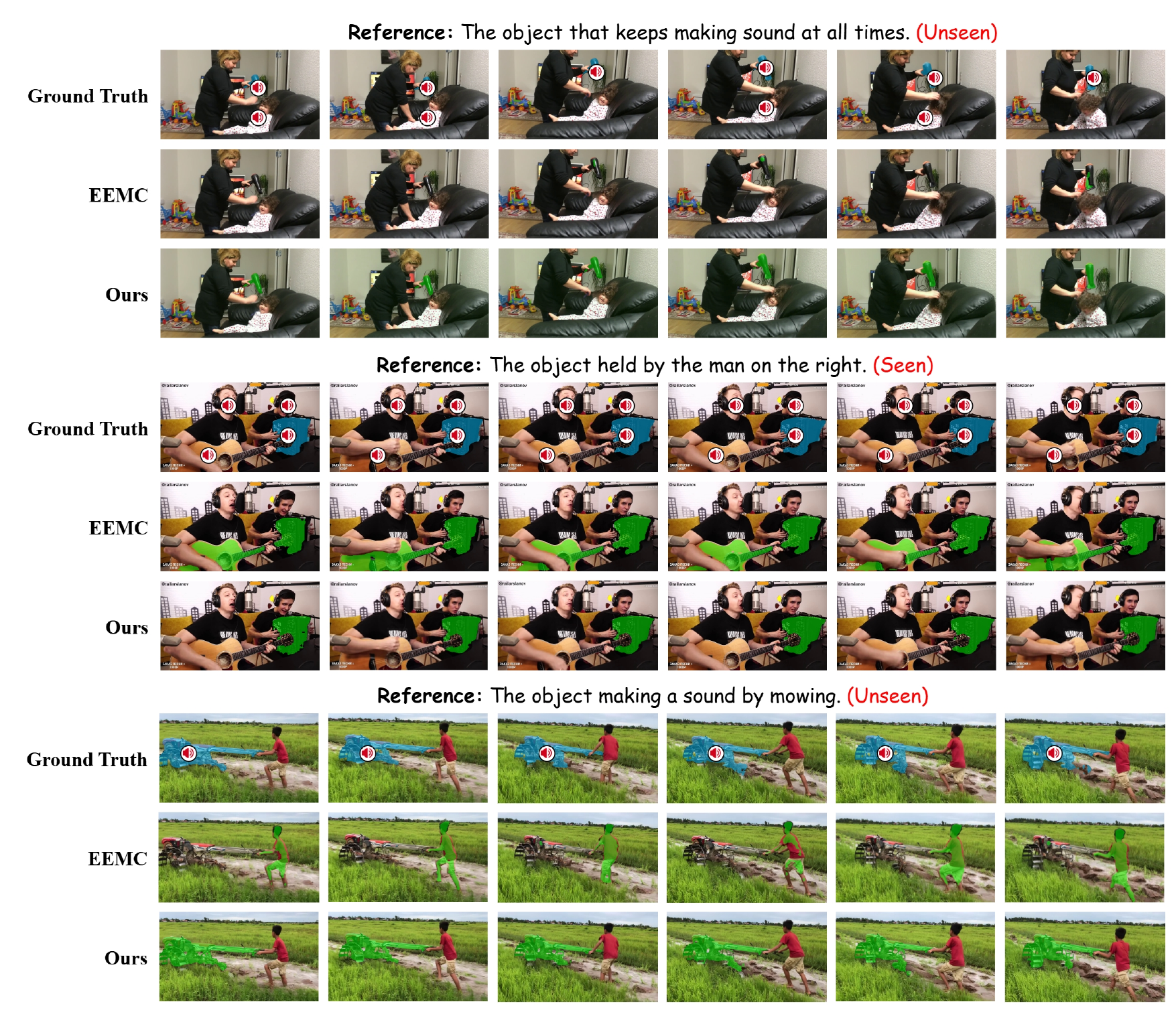}
  \caption{Visualization results of referred objects in the Ref-AVS benchmark. Please zoom in to see more details. Speaker icons indicate the sounding objects.}
  \label{fig:qualitative}
\end{figure*}
 
\subsection{Qualitative Results}
% Fig.~\ref{fig:visualization} shows the qualitative advantages of PRIMED in ambiguous audio-visual grounding scenarios. On the seen split, PRIMED produces more complete and target-aligned masks than EEMC, indicating better preservation of target integrity. On the unseen split, PRIMED more accurately localizes the correct wolf under joint ambiguity from audio and language cues, showing stronger localization accuracy in challenging unseen cases.

Cases in Fig.~\ref{fig:qualitative} illustrate the advantages of PRIMED. The first unseen case involves a small target with noticeable motion across frames, and PRIMED maintains more accurate masks than EEMC. This case suggests that PRIMED better preserves target identity for small objects in dynamic scenes. In the seen split, the expression ``\textit{the object held by the man on the right}'' mainly depends on positional and relational language cues rather than pure auditory saliency, and EEMC is more easily attracted to the two sounding instruments and fails to determine the correct target. This result indicates that the modality prior helps suppress audio distractors and bias competition toward the correct target. Another unseen case requires joint reasoning over visual and audio cues. Since both the human and the mower move prominently, but only the mower produces the sound, EEMC tends to drift toward the person, whereas PRIMED more accurately localizes the mower over time. This example highlights the role of biased competition in suppressing visually dominant but less relevant human regions.

\begin{figure*}[!t]
  \centering
  \includegraphics[width=\linewidth]{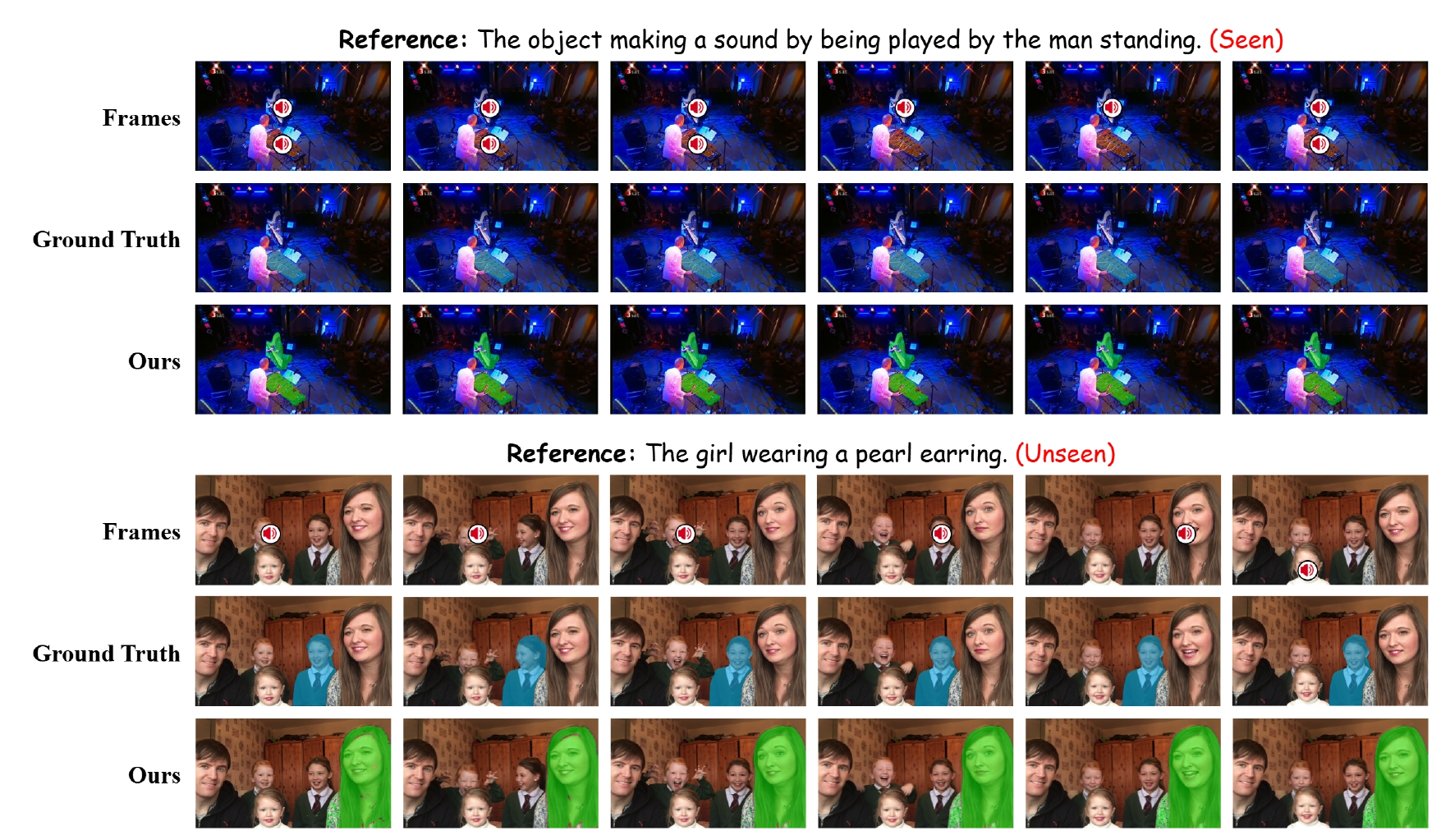}
  \caption{Failure cases of our proposed PRIMED. Please zoom in to see more details. Speaker icons indicate the sounding objects.}
  \label{fig:failure}
\end{figure*}

Failure cases are illustrated in Fig.~\ref{fig:failure}. In the first case, PRIMED tends to segment both nearby sounding objects instead of isolating the correct one. A likely reason is that the dim environment makes the human pose difficult to distinguish, resulting in unreliable visual cues for separating the correct target from adjacent candidates. This example suggests that PRIMED still has limited robustness in low-light scenes. In the second case, the main difficulty comes from fine-grained reasoning in a crowded scene with multiple human candidates. PRIMED tends to rely on salient visual evidence and fails to consistently identify the correct girl based on the subtle attribute of a ``\textit{pearl earring}''. This failure indicates that the current model remains weak when the referring expression depends on tiny discriminative details.

\subsection{Limitations and future work}
Although PRIMED achieves strong overall performance, it still shows a gap in F-score, indicating limited ability in fine boundary refinement. Moreover, compared with recent MLLM-based reasoning methods, the current framework remains less powerful in high-level semantic reasoning for complex Ref-AVS scenarios. Future work will focus on stronger boundary-aware decoding and deeper integration with MLLM-based reasoning to improve both contour quality and semantic understanding.

\section{Conclusion}
In this paper, we revisit Ref-AVS from the perspective of biased competition and present PRIMED, a SAM2-based framework that explicitly incorporates top-down modality guidance while preserving bottom-up visual evidence. By coupling the language-guided modality prior with shared global visual context, PRIMED enables more reliable target selection under multimodal ambiguity and improves segmentation robustness across challenging Ref-AVS scenarios. Extensive experiments demonstrate that each component consistently contributes to the final performance, especially in stabilizing target grounding when modality relevance varies across samples. Overall, our results suggest that explicitly modeling how different modalities participate in candidate competition offers a promising alternative to treating them as homogeneous cues, and provides a useful direction for future multimodal segmentation research.

\bibliographystyle{IEEEtran}
\bibliography{references}

\end{document}